%% file: main.tex
\documentclass{sigkddExp}

\usepackage{hyperref}
\usepackage{url}
\usepackage{xcolor}
\usepackage{footnote}
\usepackage{graphicx}
\usepackage{booktabs}
\usepackage{dsfont}

\usepackage{algorithm}
\usepackage{algorithmic}
\usepackage{subcaption}
\usepackage{multirow}

\begin{document}
%

\title{Privacy-Preserving Graph Machine Learning\\ from Data to Computation: A Survey}

\author{
Dongqi Fu\titlenote{First two authors contribute equally to this research.}$^\dagger$,~~~Wenxuan Bao$^\dagger$,~~~Ross Maciejewski$^\S$,~~~Hanghang Tong$^\dagger$,~~~Jingrui He$^\dagger$\\
\affaddr{$^\dagger$University of Illinois Urbana-Champaign}\\
\affaddr{$^\S$Arizona State University}\\
\normalsize{dongqif2@illinois.edu, wbao4@illinois.edu, rmacieje@asu.edu, htong@illinois.edu, jingrui@illinois.edu}
}


\maketitle

\begin{abstract}
In graph machine learning, data collection, sharing, and analysis often involve multiple parties, each of which may require varying levels of data security and privacy. To this end, preserving privacy is of great importance in protecting sensitive information.
In the era of big data, the relationships among data entities have become unprecedentedly complex, and more applications utilize advanced data structures (i.e., graphs) that can support network structures and relevant attribute information. To date, many graph-based AI models have been proposed (e.g., graph neural networks)  for various domain tasks, like computer vision and natural language processing.
In this paper, we focus on reviewing privacy-preserving techniques of graph machine learning. We systematically review related works from the data to the computational aspects.
We first review methods for generating privacy-preserving graph data.
Then we describe methods for transmitting privacy-preserved information (e.g., graph model parameters) to realize the optimization-based computation when data sharing among multiple parties is risky or impossible.
In addition to discussing relevant theoretical methodology and software tools, we also discuss current challenges and highlight several possible future research opportunities for privacy-preserving graph machine learning. Finally, we envision a unified and comprehensive secure graph machine learning system.
\end{abstract}

\section{Introduction}
\input{introduction}

\section{Privacy-Preserving Graph Data}
\input{graph_gen}

\section{Graph Data Privacy-Preserving Computation}
\input{graph_FL}

\section{Envisioning}
\input{envision}

\section{Conclusion}
\input{conclusion}


\section{Acknowledgements}
This work is supported by the National Science Foundation (1947203, 2117902, 2137468, 1947135, 2134079, and 1939725), the U.S. Department of Homeland Security (2017-ST-061-QA0001, 17STQAC00001-06-00, and 17STQAC00001-03-03), DARPA (HR001121C0165), NIFA (2020-67021-32799), and ARO (W911NF2110088). The views and conclusions are those of the authors and should not be interpreted as representing the official policies of the funding agencies or the government.

%
\bibliographystyle{abbrv}
\bibliography{reference}  
%
%


\end{document}

%% file: introduction.tex
According to the recent report from the United Nations~\footnote{\url{https://press.un.org/en/2022/sc15140.doc.htm}}, strengthening multilateralism is indispensable to solve the unprecedented challenges in critical areas, such as hunger crisis, misinformation, personal identity disclosure, hate speech, targeted violence, human trafficking, etc.
Addressing these problems requires collaborative efforts from governments, industry, academia, and individuals. In particular, effective and efficient data collection, sharing, and analysis are at the core of many decision-making processes, during which preserving privacy is an important topic.
Due to the distributed, sensitive, and private nature of the large volume of involved data (e.g., personally identifiable information, images, and video from surveillance cameras or body cameras), it is thus of great importance to make use of the data while avoiding the sharing and use of sensitive information.

On the other side, in the era of big data, the relationships among entities have become remarkably complicated. Graph, as a relational data structure, attracts much industrial and research interest for its carrying complex structural and attributed information. For example, with the development of graph neural networks, many application domains have obtained non-trivial improvements, such as computer vision~\cite{DBLP:journals/corr/abs-2209-13232}, natural language processing~\cite{DBLP:journals/ftml/WuCSGGLPL23}, recommender systems~\cite{DBLP:conf/ijcai/WangHW0SOC0Y21}, drug discovery~\cite{DBLP:journals/bib/GaudeletDJSRLHV21}, fraud detection~\cite{DBLP:journals/corr/abs-2106-07178}, etc.

Within the trend of applying graph machine learning methods to systematically address problems in various application domains, protecting privacy in the meanwhile is non-neglectable~\cite{fu2022privacy}. To this end, we consider two complementary strategies in this survey, namely, (1) to share faithfully generated graph data instead of the actual sensitive graph data, and (2) to enable multi-party computation without graph data sharing. Inspired by the above discussion, we focus on introducing two fundamental aspects of privacy-preserving techniques on graphs, i.e., \textbf{privacy-preserving graph data} and \textbf{graph data privacy-preserving computation}.

\begin{figure}[t]
\includegraphics[width=0.45\textwidth]{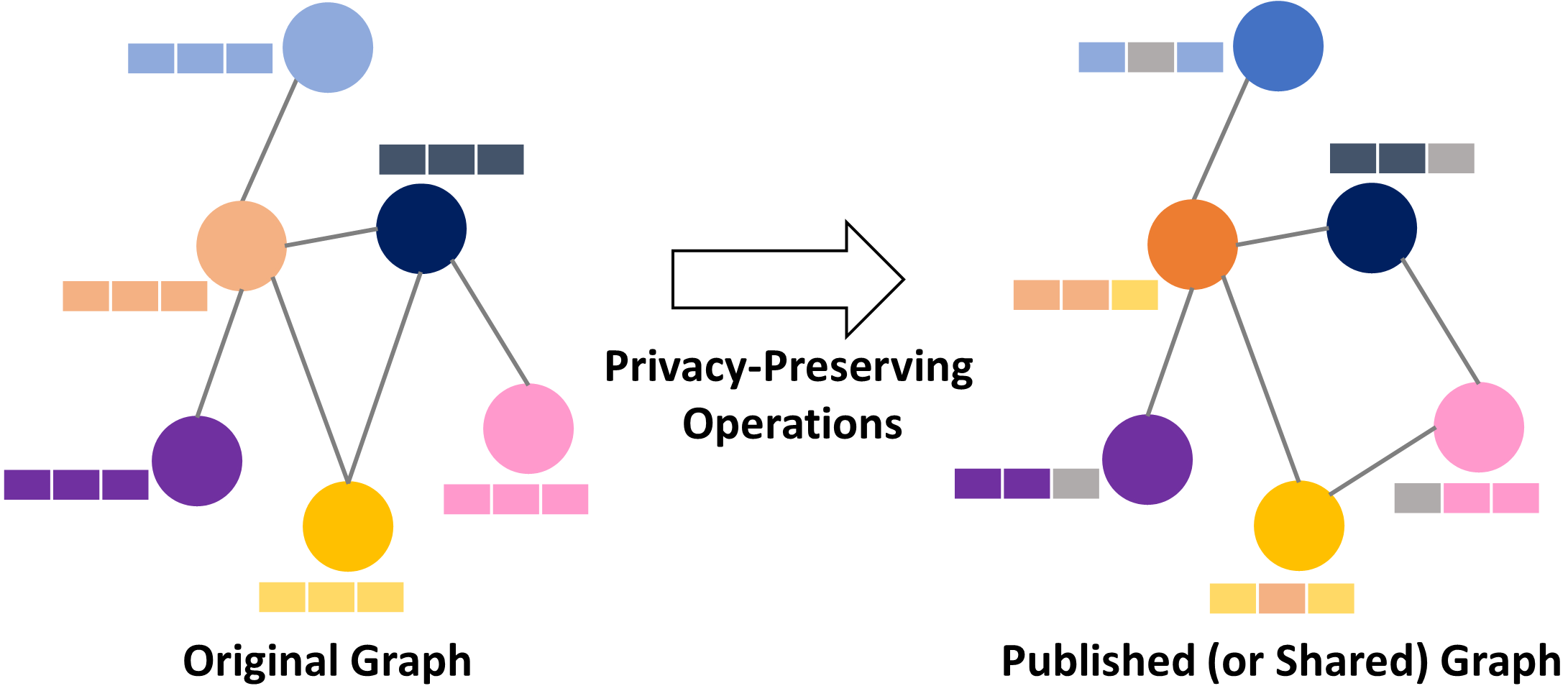}
\centering
\caption{Privacy-Preserving Graph Data. After the privacy-preserving generation, the original graph data is perturbed with certain connections and features permuted.}
\label{Fig:graph_generation}
\end{figure}

For the data aspect, \textbf{privacy-preserving graph data} as shown in Figure~\ref{Fig:graph_generation}, we focus on the scenario that when publishing or sharing the graph data is inevitable, how could we protect (e.g., mask, hide, or perturb) sensitive information in the original data to make sure that the published or shared data could survive from the external attackers (e.g., node identify disclosure and link re-identification). Hence, in Section 2, we systematically introduce various attackers~\footnote{Throughout the paper, we use ``attackers'' to denote the attacks on graphs. There are also attackers that are designed not for graphs but for Euclidean data, for example. Those are not in the scope of this paper.} first (Subsection 2.1) and what backgroud knowledge they need to execute attacks (Subsection 2.2). Then, we introduce the corresponding protection mechanisms and explain why they can address the challenges placed by attackers (Subsection 2.3). Also, we share some graph statistical properties (other than graph data itself) privacy protection mechanisms (Subsection 2.4). After that, we list several possible challenges for privacy-preserving graph data generation when facing complex structures and attributes, e.g., time-evolving graphs and heterogeneous information graphs (Subsection 2.5).

\begin{figure}[h]
\includegraphics[width=0.45\textwidth]{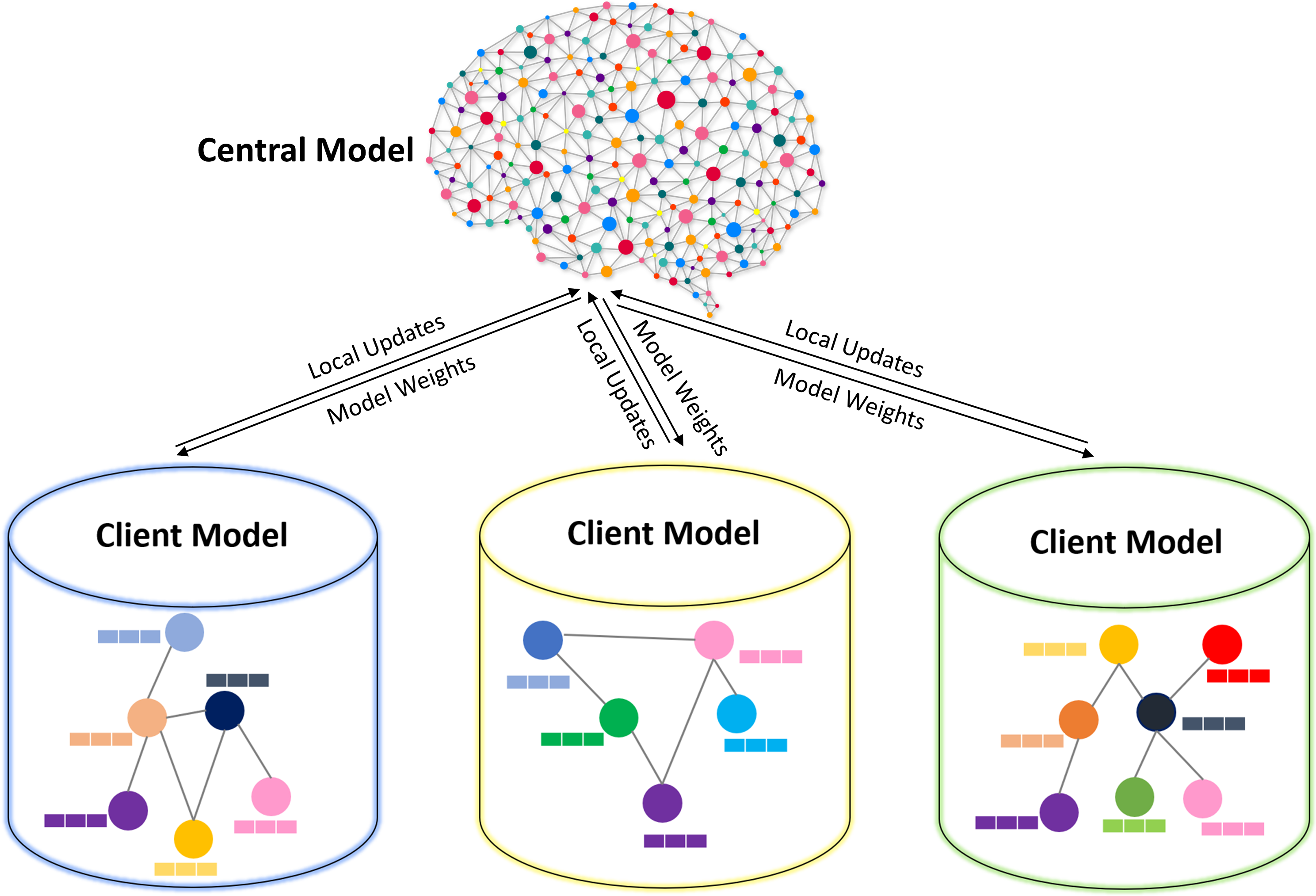}
\centering
\caption{Graph Data Privacy-Preserving Computation. In the federated learning framework, each client model has its own graph data, and the data itself is not transmitted (but the model parameters) to the central model.}
\label{Fig:graph_federated_learning}
\end{figure}

For the computation aspect, \textbf{graph data privacy-preserving computation}, we focus on the multi-party computation scenario where the input data is structured, distributed over clients, and exclusively stored (i.e., not shareable among others). Here, federated learning can be a quick-win solution. However, relational data structures (i.e., graphs) bring a significant challenge (i.e., \textit{non-IIDness}) to the traditional federated learning setting. This means that the data from intra-clients and/or inter-clients can violate the independent and identically distributed assumption (i.e., the i.i.d. assumption) due to the presence of the complex graph features, whose data complexity hinders many existing federated learning frameworks from getting the optimal performance. Motivated by this observation, in Section 3, we first discuss the adaption of federated learning on graphs and the corresponding challenge from non-IIDness brought by graphs (Subsection 3.1), then we introduce how nascent graph federated learning research works to address the non-IIDness issues from three levels, i.e., graph-level federated learning (Subsection 3.2), subgraph-level (Subsection 3.3), and node-level (Subsection 3.4). Then, we list several challenges and promising research directions, including model heterogeneity and avoiding cross-client transmission (Subsection 3.5). 

After we introduce \textbf{privacy-preserving graph data} and \textbf{graph data privacy-preserving computation} with their own methodologies, advances, software tools, limitations, and future directions. In Section 4, we envision the necessity of combing these two directions into \textbf{privacy-preserving graph data privacy-preserving computation} to meet any possibility of leaking sensitive information, to further achieve a comprehensive, well-defined, and end-to-end graph machine learning system. Finally, the paper is concluded in Section 5.

\textit{Relation with Previous Studies}.
For the \textbf{privacy-preserving graph data}, we systematically review the privacy attackers and the corresponding privacy protection techniques, which takes a balance of classic methods~\cite{DBLP:journals/sigkdd/ZhouPL08, Wu2010} and emerging solutions~\cite{DBLP:journals/tkde/JiangPYYGC23}, such as topology perturbation methods, deep generation methods, etc. Beyond that, we extend the privacy-preserving techniques review from the data level to the computation level, i.e., the \textbf{graph data privacy-preserving computation} within the federated learning framework. Most of the existing federated learning reviews do not primarily concentrate on graph federated learning~\cite{advance,fedopt,survey3,pfl_survey}. Recently, two survey papers \cite{gfl2021,gfl2022} introduce two problem settings in graph federated learning and their corresponding techniques. They exclusively focus on graph federated learning solutions and ignore the connections to traditional federated learning. Thus, we start from various application scenarios and provide a comprehensive classification and exposition of graph federated learning. While our focus primarily revolves around graph federated learning, we also highlight its connections and distinctions to traditional federated learning, aiming to present the big picture of this field. In addition to reviewing the two aspects (i.e., \textbf{privacy-preserving graph data} and \textbf{graph data privacy-preserving computation}), we also discuss the necessity and possibility of combining these two directions and propose several promising future research directions.

%% file: graph_gen.tex
\begin{figure}[h]
\includegraphics[width=0.45\textwidth]{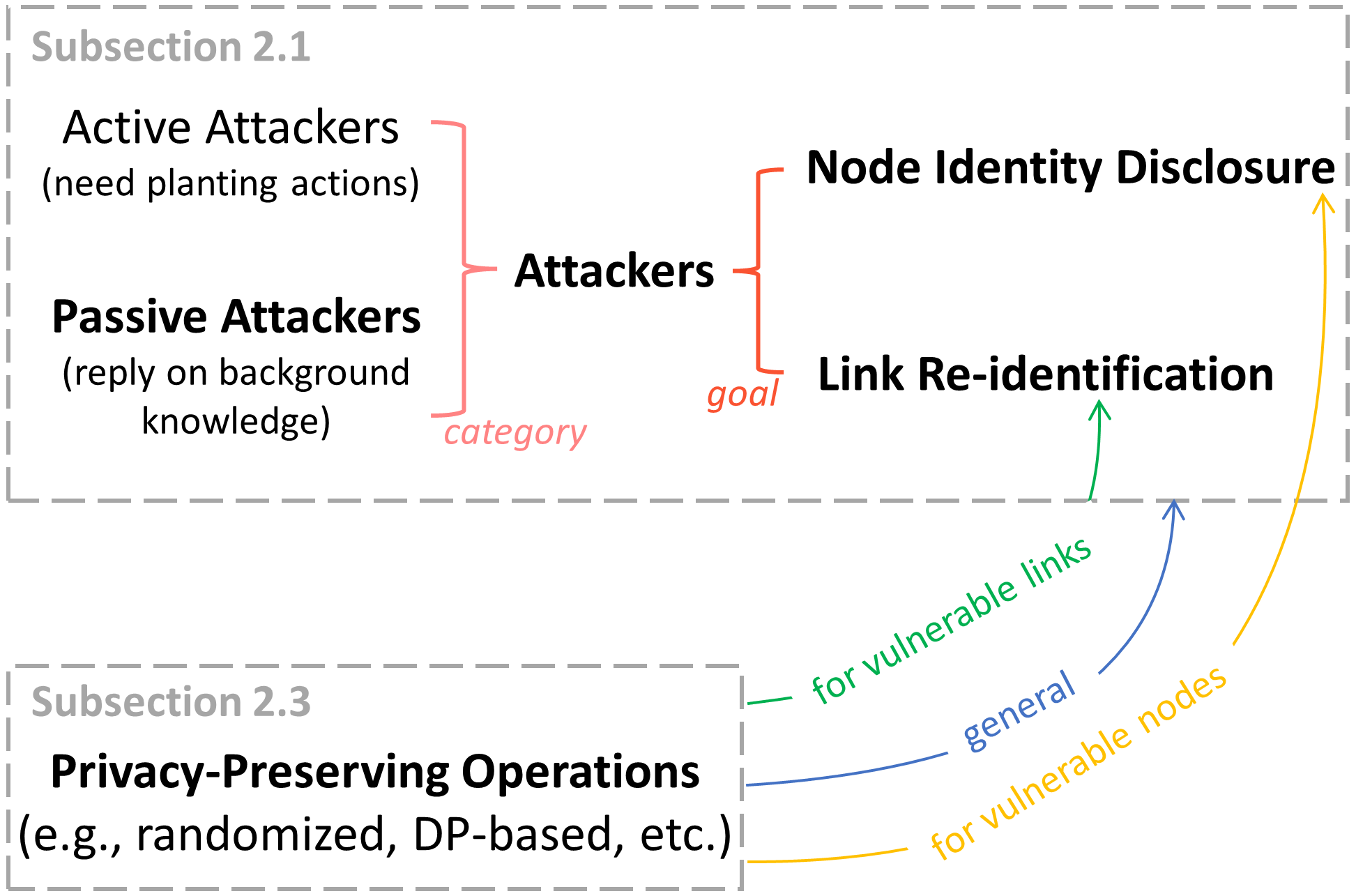}
\centering
\caption{Taxonomy Structure of Section 2.}
\label{Fig:taxo_sec_2}
\end{figure}

As for making privacy-preserving graph data to publish or share, the ultimate goal is to successfully protect the published graph data from various attacks from adversaries or attackers. To this end, we first introduce the different kinds of attackers, such as node identity disclosure or sensitive link re-identification in Subsection 2.1 and necessary background knowledge in Subsection 2.2. Then, we introduce how the corresponding privacy-preserving mechanisms are proposed, such as several of them being deliberately designed to defend against certain attackers and some of them being general protections and not aiming at specific attacks, in Subsection 2.3. The taxonomy is shown in Figure~\ref{Fig:taxo_sec_2}.

\subsection{Privacy Attackers on Graphs}
According to \cite{DBLP:conf/www/BackstromDK07}, what the attackers aim to attack is that they (1) want to learn whether edges exist or not between specific target pairs of nodes and also (2) want to reveal the true identities of targeted users, even from just a single anonymized copy of the graph, with a surprisingly small investment of effort. 

\subsubsection{Category of Attackers}
Attackers can be classified into the \textbf{active attackers} and \textbf{passive attackers}~\cite{DBLP:conf/www/BackstromDK07}.

The first category is \textbf{active attackers}, where the core idea is that the attackers actively plant certain structures into the graph before it is being published. Then, the attackers can identify victims in the published graph by locating the planted structures. For example~\cite{DBLP:journals/sigkdd/ZhouPL08}, 
the attackers create a subgraph $H$ containing $k$ nodes and then use $H$ to connect $b$ target nodes in the original graph $G$ (subgraph $H$ is better to be unique and has the property to be recovered in the published graph). After the original graph $G$ is privacy-preserved (e.g., mask and disturb connections) and published as $G'$, the attackers try to find $H$ in $G'$ and then determine those $b$ nodes.

Active attackers usually need to access the original graph beforehand and then make corresponding active actions like creating new nodes, linking new edges, and planting subgraphs. The planting and recovery operations are usually computationally costly~\cite{DBLP:conf/www/BackstromDK07}. Therefore, another direction points to passive attacks and defense.

\textbf{Passive attackers} are based on the fact or the assumption that most entities (e.g., nodes and edges) in graphs usually belong to a unique, small identifiable graph. Then, different from active attackers, passive ones do not need to create new nodes and edges in the original but mostly rely on the observation of the published graph to identify victims. In the initial proposal of passive attacks~\cite{DBLP:conf/www/BackstromDK07}, a passive attacker (e.g., a node in a social network) needs to collude with other $(k-1)$ nodes on the original graph, and the coalition needs to know the external information (e.g., their 1-hop neighbors' name in the social network), such that they can reconnect on the published graph to identify the victims. \textit{Here, we expand the scope of passive attacks to include the attackers whose core is observation plus little external information.} For example, in~\cite{DBLP:journals/pvldb/HayMJTW08}, an attacker knows the external background information like ``Greg is connected to at least two nodes, each with degree 2'' and tries to observe the candidate of plausible Greg in the published social network.

\subsubsection{Goal of Attackers}
The ultimate goals of most graph privacy attackers can be roughly divided into disclosing the node identity (e.g., name, DOB, and SSN in the social network) and the link existence (e.g., sensitive connections in the social network)~\cite{DBLP:books/tf/08/0001DGK08, Wu2010, DBLP:journals/kais/YingW11, DBLP:journals/sigkdd/ZhouPL08}. Next, we formally introduce the general definition of these two goals.

\textbf{Node Identity Disclosure}.
The node identity disclosure problem often arises from the scenario that the attackers aim to identify a target node identity in the published graph (usually, which has been anonymized already). For example, in a published social network with usernames masked already, the node identity disclosure aims to identify which node is Greg~\cite{DBLP:journals/pvldb/HayMJTW08}. To be more specific, the identity disclosure can be detailedly divided into node existence disclosure (i.e., whether  a target node existed or not in a published graph), node property disclosure (i.e., partial features of a target node are disclosed like its degree, distance to the center, or even sensitive labels, etc)~\cite{DBLP:journals/sigkdd/ZhouPL08}.

\textbf{Link Re-Identification}.
In a given graph, edges may be of different types and can be classified as either sensitive or not. Some links (i.e., edges) are safe to release to the public, such as classmates or friendships. And some links are sensitive and should maintain private but not published, like the personal disease records with hospitals. The problem of link re-identified is defined as inferring or predicting sensitive relationships from anonymized graphs~\cite{DBLP:conf/kdd/ZhelevaG07}. Briefly speaking, the adversary (or attacker) achieves the goal when it is able to correctly predict a sensitive link between two nodes. For example, if the attacker can figure out which there is a transaction between two users, given the properties of the released financial graph. Also, there are some detailed categorizations of the line re-identification other than the link existence, such as the link weight and link type or labels~\cite{DBLP:journals/sigkdd/ZhouPL08}.

Compared with active attackers, passive attackers are typically efficient in executing for adversaries and do not need to interact with the original graph beforehand very much. \textbf{Thus, within the scope of passive attackers, achieving those attacking goals (node identity disclosure or link re-identification) relies on the observation of the published graph and certain external background knowledge to further identify victims.\footnote{Node identity disclosure and link re-identification can also be achieved in active ways~\cite{DBLP:conf/www/BackstromDK07}, but in the paper, we focus on introducing the passive manners that achieve those goals.}} Next, we focus on introducing what requirements passive attackers need to execute attacks passively.

\subsection{Background Knowledge for Passive Attacks}
Here, we first discuss some background knowledge that could contribute to the goal of node identity disclosure. Then, we list some background knowledge that could contribute to sensitive link re-identification attacks.

\subsubsection{Background Knowledge for Node Identity Disclosure}

In general, the background knowledge for achieving node identity disclosure is to help them to detect the uniqueness of victims (i.e., nodes in the published graph) and thus narrow down the scope of candidate sets to increase the successful attack probability. For example, assume that the attackers know some background knowledge $\mathcal{H}$ about a target node, after that, the attackers observe the published graph and find 2 candidates satisfying the condition (i.e., $\mathcal{H}$), then the attackers have $50\%$ confidence to reveal the identity of that target node in the published graph. Next, we introduce some methods to acquire background knowledge.

\textbf{Vertex Refinement Queries}~\cite{DBLP:journals/pvldb/HayMJTW08}. These are interactive queries, which describe the local structure of the graph around a target node $x$. The initial query in vertex refinement queries is denoted as $\mathcal{H}_{0}(x)$ that simply returns the label of node $x$ in the labeled graph (or a constant $\epsilon$ in the unlabeled graph). And $\mathcal{H}_{1}(x)$ returns the degree of node $x$. Then, iteratively, $\mathcal{H}_{i}(x)$ is defined as the multiset of $\mathcal{H}_{i-1}(\cdot)$ queries on 1-hop neighbors of node $x$, which can be expressed as follows.
\begin{equation}
    \mathcal{H}_{i}(x) = \{\mathcal{H}_{i-1}(z_{1}), \mathcal{H}_{i-1}(z_{2}), \ldots, \mathcal{H}_{i-1}(z_{d_{x}})\}
\end{equation}
where $d_{x}$ is the degree of node $x$. For example, in a social network, $\mathcal{H}_{2}(Bob)=\{1,1,4,4\}$ means that Bob has four neighbors their degrees are 1, 1, 4, and 4, respectively.

\textbf{Subgraph Queries}~\cite{DBLP:journals/pvldb/HayMJTW08}. These queries assert the existence of a subgraph around a target node. Compared with the above vertex refinement queries, subgraph queries are more general (i.e., the information is not exclusively occupied to a certain graph structure) and flexible (i.e., informativeness is not limited by the degree of a target node). In brief, the adversary is assumed capable of gathering some fixed number of edges around a target node $x$ and figuring out what subgraph structure those collected edges can form. For example, still targeting Bob in a social network, when collecting 3 edges, attackers can find 3 distinct neighbors. And collecting 4 edges can find a tree rooted by Bob. Those existences of structures form $H$ such that attackers can use them to reveal the identity of Bob. Also, different searching strategies can result in different subgraph structures. For example, based on collecting 3 edges from Bob, breadth-first exploration may result in a star subgraph, and depth-first exploration may end up with a three-node-line. We refer to~\cite{DBLP:journals/pvldb/HayMJTW08}, where a range of searching strategies are tested to empirically illustrate the descriptive power of background knowledge.

\textbf{Hub Fingerprint Queries}~\cite{DBLP:journals/pvldb/HayMJTW08}. First of all, a hub stands for a node that has a high degree and a high betweenness centrality (i.e., the proportion of shortest paths in the graph that include that node) in the graph. Then, a hub fingerprint is the description of a node's connections to hubs. To be more specific, for a target node $x$, the corresponding hub fingerprint query $\mathcal{H}_{i}(x)$ records the shortest distance towards each hub in a graph. In $\mathcal{H}_{i}(x)$, $i$ is the limit of measurable distance. For example, $\mathcal{H}_{1}(Bob) = (1,0)$ means Bob is 1 distance away from the first hop and not connected to (or 1 distance non-reachable from) the second hub. And, $\mathcal{H}_{2}(Bob) = (1,2)$ means that Bob is 1 distance away from the first hop and 2 distance away from the second hub.

\textbf{Neighborhood Relationships Queries}~\cite{DBLP:conf/icde/ZhouP08}. Targeting a node, if an adversary has background knowledge about its neighbors and the relationship among the neighbors, then the victim can be identified in the anonymized graph. To be specific, the neighborhood relationship query rely more on the isomorphism of the ego-graph (i.e., 1-hop neighbors) of a target node to reveal its identity, compared with iterative vertex refinement query~\cite{DBLP:journals/pvldb/HayMJTW08} and general subgraph query~\cite{DBLP:journals/pvldb/HayMJTW08}. For example, in a social network, if Bob has two close friends who know each other (i.e., are connected) and two close friends who do not know each other (i.e., are not connected), then this unique information obtained by the adversary can be used to find Bob in the published anonymized graph.

\subsubsection{Background Knowledge for Link Re-Identification}
\textbf{Link Prediction Probabilistic Model}~\cite{DBLP:conf/kdd/ZhelevaG07}. This probabilistic model is proposed to determine whether a relationship between two target nodes. And different kinds of background information (i.e., observation) can be leveraged to formalize the probabilistic model, such as (1) \textit{node attributes}, e.g., two social network users who share the same interest are more likely to be friends; (2) \textit{existing relationships}, e.g., two social network users in the same community are more likely to be friends; (3) \textit{structural properties}, e.g., the high degree nodes are more likely to connect in a graph; and (4) inferred relationships (i.e., a complex observation that is more likely based on the inference of the invisible relationship), e.g., two social network users are more likely to be friends if they both are close friends of a third user. 

Mathematically, those above observations can be expressed for predicting the existence of a sensitive relation between node $i$ and node $j$ as $P(e^{s}_{ij}|O)$, where $e^{s}_{ij}$ stands for the sensitive relationship and $O$ consists of several observations $\{o_{1}$, \ldots , $o_{n}\}$. For example, if we use the second kind of information (i.e., existing relationships), then $\{o_{1}$, \ldots , $o_{n}\}$ is a set of edges between node $i$ and node $j$ with the edge type other than $s$, denoted as $e^{l}_{ij}$ and $l \in \{1, \ldots, n\}$ is the index of other edge relationships. To solve out $P(e^{s}_{ij}|O)$, the noisy-or model~\cite{pearl1988probabilistic} can be used as suggested by ~\cite{DBLP:conf/kdd/ZhelevaG07}, where each observation $o_{l} \in \{o_{1}, \ldots , o_{n}\}$ is considered as independent with each other and parameterised as $\lambda_{l} \in \{\lambda_{1}, \ldots , \lambda_{n}\}$. Moreover, there is a leak parameter $\lambda_{0}$ to capture the probability that the sensitive edge is there due to other unmodeled reasons. Hence, the probability of a sensitive edge is expressed as follows.
\begin{equation}
    P(e^{s}_{ij} = 1 | o_{1}, \ldots, o_{n}) = 1 - \prod_{l=0}^{n}(1- \lambda_{l})
\end{equation}
where $s$ in $e^{s}_{ij}$ is the indicator of \underline{s}ensitive relationship, and the details of fitting the values of $ \lambda_{l}$ can be found in~\cite{DBLP:conf/kdd/ZhelevaG07}.

\textbf{Randomization-based Posterior Probability}~\cite{DBLP:conf/sdm/YingW09a}. To identify a link, this observation is based on randomizing the published graph $G'$ and counting the possible connections over a target pair of nodes $i$ and $j$. And those countings are utilized for the posterior probability to determine whether there is a link between nodes $i$ and $j$ in the original graph $G$. Formally, the posterior probability for identifying the link $e_{ij}$ in the original graph $G$ is expressed as follows.
\begin{equation}
    P(e_{ij} = 1 | G'_{s}) = \frac{1}{N} \sum^{N}_{s=1} \mathds{1} (G'_{s}(i,j) == 1)
\end{equation}
where the attacker applies a certain randomization mechanism on the published graph $G'$ $N$ times to get a sequence of $G'_{s}$, and $s \in \{1, \ldots, N\}$. In each $G'_{s}$, if there is an edge connects the target nodes $i$ and $j$, then the indicator function $\mathds{1} (G'_{s}(i,j) == 1)$ will count one.

\subsection{Privacy-Preserving Mechanisms}
Here, we discuss some privacy-preserving techniques that are deliberately designed for specific attackers and also some general protection techniques that are not targeting attackers but can be widely applied.

\subsubsection{Protection Mechanism Designed for Node Identity Dislosure}
In general, the protection mechanisms are proposed to enlarge the scope of candidates of victims, i.e., reduce the uniqueness of victims in the anonymized graphs.

\textbf{$k$-degree Anonymization}~\cite{DBLP:conf/sigmod/LiuT08}. The motivation for $k$-degree anonymization is that degree distribution is highly skewed in real-world graphs, such that it is usually effective to collect the degree information (as the background knowledge) to identify a target node. Therefore, this protection mechanism aims to ensure that there at least exist $k-1$ nodes in the published graph $G'$, in which $k-1$ nodes share the same degree with any possible target node $x$. In this way, it can largely prevent the node identity disclosure even if the adversary has some background knowledge about degree distribution. To obtain such anonymized graph $G'$, the method is two-step. First, for the original graph $G$ with $n$ nodes, the degree distribution is encoded into a $n$-dimensional vector $\mathbf{d}$, where each entry records the degree of an individual node; And then, based on $\mathbf{d}$, the authors proposed to create a new degree distribution $\mathbf{d}'$, which is $k$-anonymous with a tolerated utility loss (e.g., isomorphism cost) instanced by the $L_{1}$ distance between two vectors $\mathbf{d}$ and $\mathbf{d}'$. Second, based on the $k$-anonymous degree vector $\mathbf{d}'$, the authors proposed to construct a graph $G'$ whose degree distribution is identical to $\mathbf{d}'$.

\textbf{$k$-degree Anonymization in Temporal Graphs}~\cite{DBLP:conf/icwsm/0004MT15}. For temporal graphs (i.e., graph structures and attributes are dependent on time~\cite{fu2022natural}), this method aims to ensure that the temporal degree sequence of each node is indistinguishable from that of at least $k-1$ other nodes. On the other side, this method also tries to preserve the utility of the published graph as much as possible. To achieve the $k$-anonymity, the proposed method first partition $n$ nodes in the original temporal graph $G$ into $m$ groups using $k$-means based on the distance of temporal degree vectors $\mathbf{d}$ of each node, which is a $T$-dimensional vector records the degree of a node at different timestamp $t$.To realize the utility, constrained by the cluster assignment, the method refines $\mathbf{d}$ of each node into $\mathbf{d}'$ while minimizing the $L_{1}$ distance between matrices $\mathbf{D}$ and $\mathbf{D}'$ (which are stacks of $\mathbf{d}$ and $\mathbf{d}'$). After that, the anonymized temporal graph $G'$ is constructed by $\mathbf{D}'$ to release for each timestamp individually.

\textbf{$k$-degree Anonymization in Knowledge Graphs}~\cite{DBLP:conf/icde/HoangCF21}. Different from the ordinary graph, the knowledge graph has rich attributes on nodes and edges~\cite{DBLP:journals/tnn/JiPCMY22}. Therefore, the $k$-degree is upgraded with the $k$-attributed degree that aims to ensure a target node in the anonymized knowledge graph has $k-1$ other nodes who share the same attributes (i.e., node level) and degree (i.e., edge level)~\cite{DBLP:conf/acns/HoangCF20}. Then the $k$-degree anonymization solution gets upgraded in~\cite{DBLP:conf/icde/HoangCF21}, which aims to solve the challenge when the data provider wants to continually publish a sequence of anonymized knowledge graphs (e.g., the original graph needs to update and so the anonymized does). Then, in~\cite{DBLP:conf/icde/HoangCF21}, the $k$-ad (short for $k$-attributed degree) is extended to $k^{\omega}$-ad, which targets to defend the node identity disclosure in the $\omega$ continuous anonymized versions of a knowledge graph. The basic idea is to partition nodes into clusters based on the similarity of node features and degree; Then, for the knowledge graph updates (like newly inserted nodes or deleted nodes), manual intervention is applied (e.g., adding fake nodes) to ensure the $k^{\omega}$ anonymity; Finally, the anonymized knowledge graph gets recovered from the clusters. This initial idea~\cite{DBLP:conf/icde/HoangCF21} gets further formalized and materialized in~\cite{hoang2022time}.

\textbf{$k$-neighborhood Anonymization}~\cite{DBLP:conf/icde/ZhouP08}. This protection is proposed to defend the node identity disclosure when the adversary comprehends the background knowledge about neighborhood relationships of a target node (i.e., Neighborhood Relationship Queries discussed in Subsection 2.2.1). The core idea is to insert nodes and edges in the original graph $G$ to get an anonymized graph $G'$, such that a target node $x$ can have multiple nodes whose neighborhood structure is isomorphic in $G'$. Given a pair node $v$ and $u$ in graph $G$ (suppose node $v$ is the target), the authors first propose the \textit{neighborhood component} and use DFS search to encode the ego-net $Neighbor_{G}(v)$ and $Neighbor_{G}(u)$ into vectors. Then, by comparing the difference between $Neighbor_{G}(v)$ and $Neighbor_{G}(u)$, the authors then greedy insert missing (labeled) nodes and edges (into $Neighbor_{G}(v)$ or $Neighbor_{G}(u)$) to make $Neighbor_{G}(v)$ and $Neighbor_{G}(u)$ isomorphic. Those inserted nodes and edges make $G$ into $G'$.

\textbf{$k$-automorphism Anonymization}~\cite{DBLP:journals/pvldb/ZouCO09a}. This method is proposed for the structural queries by attackers, especially for the subgraph queries (as discussed in Subsection 2.2.1). Basically, given an original graph $G$, this method produces an anonymization graph $G'$ to publish, where $G$ is the sub-graph of $G'$ and $G'$ is k-automorphic. To do this, the authors propose the KM algorithm, which partitions the original graph $G$ and adds the crossing edge copies into $G$, to further convert $G$ into $G'$. Hence, the $G'$ can satisfy the $k$-different match principle to defend the subgraph query attacks, which means that there are at least $k$-different matches in $G'$ for a subgraph query, but those matches do not share any nodes.

\subsubsection{Protection Mechanism Designed for Link Re-Identification}
The general idea of solutions here is proposed to reduce the confidence of attackers (which usually can be realized by a probabilistic model) for inferring or predicting links based on observing the published anonymized graphs.

\textbf{Intact Edges}~\cite{DBLP:conf/kdd/ZhelevaG07}. This solution is straightforward and trivial. Given the link re-identification attacker aims to predict a target link between two nodes, and the corresponding link type (i.e., edge type) is denoted as $s$, then the intact edges strategy is to remove all $s$ type edges in the original graph $G$ and publish the rest as the anonymized graph $G'$. Those remaining edges are so-called intact.

\textbf{Partial-edge Removal}~\cite{DBLP:conf/kdd/ZhelevaG07}. This approach is also based on removing edges in the original graph $G$ to publish the anonymized graph $G'$. Partial-edge removal does not exhaustively remove all sensitive (indexed by $s$ type) edges in $G$, but it removes part of existing edges. Those removed existing edges are selected based on the criteria of whether their existence contributes to the exposure of sensitive links, e.g., they are sensitive edges, they connect high-degree nodes, etc. Even those removals can be selected randomly.

\textbf{Cluster-edge Anonymization}~\cite{DBLP:conf/kdd/ZhelevaG07}. This method requires that the original graph $G$ can be partitioned into clusters (or so-called equivalence classes) to publish the anomymized graph $G'$. The intra-cluster edges are removed to aggregate a cluster into a supernode (i.e., the number of clusters in $G$ is now the number of nodes in $G'$), but the inter-cluster edges are reserved in $G'$. To be more specific, for each edge whose edge type is not sensitive (i.e., not $s$ type), if it connects any two clusters, it will be reserved in $G'$; otherwise, it will be removed. It can be observed that this method needs the clustering pre-processing, which also means that it can cooperate with the node anonymization method. For example, the $k$-anonymization~\cite{DBLP:journals/tkde/Samarati01,DBLP:conf/icde/LiLV07, DBLP:journals/tkdd/MachanavajjhalaKGV07} can be applied on the original graph $G$ first to identify the equivalence classes, i.e., which nodes are equivalent in terms of $k$-anonymization (for example, nodes who have the same degree).

\textbf{Cluster-edge Anonymization with Constraints}~\cite{DBLP:conf/kdd/ZhelevaG07}. This method is the upgraded version of the previous cluster-edge anonymization, and it is proposed to strengthen the utility of the anonymized graph $G'$ by adjusting the edges between clusters (i.e., equivalence classes). The core idea is to require the equivalence class nodes (i.e., cluster nodes or supernodes in $G'$) to have the same constraints as any two nodes in the original graph $G$. For example, if there can be at most two edges of a certain type between nodes in $G$, there can be at most two edges of a certain type between the cluster nodes in $G'$.

\subsubsection{General Privacy Protection Mechanisms}
Besides the protections that are designed deliberately for the node identity disclosure and link re-identification risks, there are also other protection mechanisms that are not designed for a specific kind of attacker but for the general and comprehensive scenario, such as randomized mechanisms with constraints and differential privacy schema. Next, we will discuss these research works.

\textbf{Graph Summarization}~\cite{DBLP:journals/pvldb/HayMJTW08}. This method aims to publish a set of anonymized graphs $G'$ given an original graph $G$, through the graph summarization manner. To be specific, this method relies on a pre-defined partitioning method to partition the original graph $G$ into several clusters, then each cluster will just serve as a node in the anonymized graph $G'$. The selection of connecting nodes in $G'$ results in the variety of $G'$, which means that a sequence of $G'$ will appear with a different edge connecting strategy. The detailed connection strategy can be referred to ~\cite{DBLP:journals/pvldb/HayMJTW08}.

\textbf{Switching-based Graph Generation}~\cite{DBLP:conf/sdm/YingW09a}. Here, the authors aim to publish the anonymized graph $G'$ that should also preserve the utility of the original graph $G$. Therefore, they propose the graph generation method based on the switching operations that can preserve the graph features. Moreover, the switching is realized in an iterative Monte Carlo manner, each time two edges ($a$, $b$) and ($c$, $d$) are selected. Then they will switch into ($a$, $d$) and ($b$, $c$) or ($a$, $c$) and ($b$, $d$). The authors constrain that two selected edges are switchable if and only if the switching generates no more edges or self-edges, such that the overall degree distribution will not change. After sufficient Monte Carlo switching operations, the authors show that the original graph features (e.g., eigenvalues of adjacency matrix, eigenvectors of Laplacian matrix, harmonic mean of geodesic path, and graph transitivity) can be largely preserved in the anonymized graph $G'$.

\textbf{Spectral Add/Del and Spectral Switch}~\cite{DBLP:conf/sdm/YangW08}. The idea of this method starts from Rand Add/Del and Rand Switch. Rand Add/Del means that the protection mechanism randomly adds an edge after deleting another edge and repeats multiple times, such that the total number of edges in the anonymized graph will not change. Rand Switch is the method that randomly switches a pair of existing edges ($t,w$) and ($u$, $v$) into $(t,v)$ and $(u,w)$ (if $(t,v)$ and $(u,w)$ do not exist in the original graph), such that the overall degree distribution will not change. In~\cite{DBLP:conf/sdm/YangW08}, the authors develop the spectrum-preserving randomization methods Spectral Add/Del and Spectral Switch, which preserve the largest eigenvalue $\lambda_{1}$ of the adjacency matrix $\mathbf{A}$ and the second smallest eigenvalue $\mu_{2}$ of the Laplacian matrix $\mathbf{L} = \mathbf{D} - \mathbf{A}$. To be specific, the authors first investigate which edges will cause the $\lambda_{1}$ and $\mu_{2}$ increase or decrease in the anonymized graph and then select the edges from different categories to do Rand Add/Del and Rand Switch to control the values of $\lambda_{1}$ and $\mu_{2}$ not change too much in the anonymized graph.

\textbf{RandWalk-Mod}~\cite{DBLP:conf/ccs/NguyenIR15}. This method aims to inject the connection uncertainty by iteratively copying each existing edge from the original graph $G$ to an initial null graph $G'$ with a certain probability, guaranteeing the degree distribution of $G'$ is unchanged compared with $G$. Starting from each node $u$ in the original graph $G$, this method first gets the neighbor of node $u$ in $G$ denoted as $\mathcal{N}_{u}$. Then for each node in $\mathcal{N}_{u}$, this method runs multiple random walks and denotes the terminated node in each walk as $z$. Finally, RandWalk-Mod adds the edge $(u,z)$ to $G'$ with certain probabilities under different conditions (e.g., 0.5, a predefined probability $\alpha$, or $\frac{0.5d_{u} - \alpha}{d_{u}-1}$, where $d_{u}$ is the degree of node $u$ in $G$).

\begin{table*}[t]
\caption{Graph Privacy-Preserving Mechanisms}
\centering
\footnotesize
\begin{tabular}{lp{4.3cm}p{6.8cm}p{1.6cm}}

\toprule 
Scenario & Name & Description & Link \\
\midrule

\multirow{6}{*}{Node Identity Disclosure} 
& $k$-degree Anonymization~\cite{DBLP:conf/sigmod/LiuT08} &  Generate $k-1$ plausible candidate for protecting the victim & \href{https://github.com/blextar/graph-k-degree-anonymity}{Github (Unofficial)}  \\
& $k$-degree Anonymization in Temporal Graphs~\cite{DBLP:conf/icwsm/0004MT15} &  Adaption of $k$-degree Anonymization on temporal graphs & ------  \\
& $k$-degree Anonymization in Knowledge Graphs~\cite{hoang2022time} &  Adaption of $k$-degree Anonymization on knowledge graphs & \href{https://github.com/tuhoag/personalized-anony-kg}{Github}  \\

\midrule

\multirow{3}{*}{Link Re-identification} 
& Partial-edge Removal, Cluster-edge Anonymization~\cite{DBLP:conf/kdd/ZhelevaG07} &    Edge removing methods that are deliberately designed for link re-identification tasks & ------\\

\midrule

\multirow{11}{*}{General} 
& Graph Summarization \cite{DBLP:journals/pvldb/HayMJTW08} &  Partitioning (or clustering) + Publishing supernodes and superedges  & \href{https://github.com/michaelghay/graph-gen}{Github (Unofficial)} \\

& Local Differential Privacy Graph Generation~\cite{DBLP:conf/ccs/QinYYKX017} &  Proportionally flipping existing and non-existing edges with graph utility preserved  & ------ \\

& Edge-level DP Algorithm~\cite{DBLP:conf/ijcai/YangWZCS21} &  A deep learning graph generative model under the edge-level differential privacy constraints  & \href{https://github.com/haonan3/Secure-Network-Release-with-Link-Privacy}{Github} \\

& Node-level DP Algorithms \cite{DBLP:conf/tcc/KasiviswanathanNRS13} & Some node-level differential privacy algorithms that compute low-sensitivity approximations to several classes of graph statistics  & \href{https://github.com/anusii/graph-dp}{Github (Unofficial)} \\

\bottomrule
\end{tabular}
\end{table*}

Next, we introduce an important component in the graph privacy-preserving techniques, i.e., differential privacy~\cite{DBLP:journals/tkde/JiangPYYGC23}. The general idea of differential privacy is that two adjacent graphs (e.g., one node/edge difference between two graphs) are indistinguishable through the permutation algorithm $\mathcal{M}$. Then, this permutation algorithm $\mathcal{M}$ satisfies the differential privacy. The behind intuition is that the randomness of $\mathcal{M}$ will not make the small divergence produce a considerably different distribution, i.e., the randomness of $\mathcal{M}$ is not the cause of the privacy leak. If the indistinguishable property is measured by $\epsilon$, then the algorithm is usually called $\epsilon$-differential privacy algorithm. The basic idea can be expressed as follows.
\begin{equation}
    \frac{Pr[\mathcal{M}(G) \in S]}{Pr[\mathcal{M}(\Tilde{G}) \in S]} \leq e^{\epsilon}
\end{equation}
where $G$ and $G'$ are adjacenct graphs, $\mathcal{M}$ is the differential privacy algorithm, and $\epsilon$ is the privacy budget. The above equation illustrates that the probability of the same output range is almost equivalent.

Within the context of graph privacy, the differential privacy algorithm can be roughly categorized as edge-level differential privacy and node-level differential privacy. Given the input original graph $G$, the output graph of the differential algorithm $\mathcal{M}(G)$ can be used as the anonymized graph $G'$ to publish.

\textbf{Edge-level Differential Privacy Graph Generation}. We first introduce the edge-level differential privacy algorithms, which means that the privacy algorithm can permute adjacent graphs (e.g., one edge difference) indiscriminately.
\begin{itemize}
    \item \textbf{DP-1K and DP-2K Graph Model}~\cite{DBLP:journals/tdp/0009W13}. This edge-level differential privacy algorithm is proposed with the utility preserving concern of complex degree distribution. Here, 1k-distribution denoted by $P_{1}(G)$ is the ordinary node degree distribution in graph $G$, e.g., the number of nodes having 1 degree is 10 then $P_{1}(1) = 10$, the number of nodes having 2 degrees is 5 then $P_{1}(2) = 5$, etc. 2K-distribution denoted by $P_{2}(G)$ is the joint graph distribution in graph $G$, e.g., the number of edges connecting an $i$-degree node and a $j$-degree node, with iterating $i$ and $j$. And $P_{2}(2,3) = 6$ means that the number of edges in $G$ connecting a 2-degree node and a 3-degree node is 6. Hence, DP-1K (or DP-2K) Graph Model first computes the 1K-(or 2K-) degree distribution $P_{1}(G)$ (or $P_{2}(G)$) and then permutes the degree distribution under the edge-level DP to obtain the $P_{1}(G)'$ (or $P_{2}(G)'$). Finally, an off-the-shelf graph generator (e.g., \cite{DBLP:conf/imc/SalaZWZZ11}) is called to build the anonymized graph $G'$ based on $P_{1}(G)'$ (or $P_{2}(G)'$).
    \item \textbf{Local Differential Privacy Graph Generation} (LDPGEN)~\cite{DBLP:conf/ccs/QinYYKX017} is motivated by permuting the connection distribution, i.e., proportionally flipping the existing edge to non-existing and vice versa. To make the generated graph preserve the original utility, LDPGEN~\cite{DBLP:conf/ccs/QinYYKX017} first partitions the original graph $G$ into the disjoint clusters and adds Laplacian noise on the node's degree vector in each cluster, which guarantees the local edge-level differential privacy. After that, the estimator is used to estimate the connection probability of intra-cluster edges and inter-cluster edges based on the noisy degree vectors, such that the anonymized graph $G'$ is generated.
    \item \textbf{Differentially Private Graph Sparsification}~\cite{DBLP:conf/nips/AroraU19}. On the one hand, this method constrains the number of edges in the anonymized graph $G'$ is less than the original graph $G$ to a certain extent. On the other hand, the method requires that the Laplacian of the anonymized graph $G'$ is approximated to the original graph $G$ (i.e., see Eq.1 in~\cite{DBLP:conf/nips/AroraU19}). The two above objectives are unified into an edge-level differential privacy framework. The new graph $G'$ is then obtained by solving an SDP (i.e., semi-definite program) problem.
    \item \textbf{Temporal Edge-level Differential Privacy}. In~\cite{DBLP:conf/aistats/UpadhyayUA21}, two temporal graphs are adjacent if they only differ in one update (i.e., the existence and non-existence of a temporal edge, different weights of an existing temporal edge). Based on the Priv-Graph algorithm (i.e., adding noise to graph Laplacian matrix), Sliding-Priv-Graph~\cite{DBLP:conf/aistats/UpadhyayUA21} is proposed to (1) take recent updates and ensure the temporal edge-level differential privacy and (2) meet the smooth Laplacian property (i.e., the positive semi-definite of consecutive Laplacian matrices). Moreover, in~\cite{DBLP:conf/esa/FichtenbergerHO21}, the authors distinguish the edge-adjacency and node-adjacency in the temporal graphs. Two temporal graphs are node-adjacent (or edge-adjacent) if they only differ in one node (or edge) insertion or deletion.
    \item \textbf{Deep Graph Models with Differential Privacy}. Following the synergy of deep learning and differential privacy~\cite{DBLP:conf/ccs/AbadiCGMMT016}, another way to preserve privacy is targeting the gradient of deep graph learning models. In~\cite{DBLP:conf/ijcai/YangWZCS21}, a deep graph generative model called $DPGG_{AN}$ is proposed under the edge-level differential privacy constraints, where the privacy protection mechanism is executed during the gradient descent phase of the generation learning process, by adding Gaussian noise to the gradient of deep learning models.
\end{itemize}

\textbf{Node-level Differential Privacy Graph Generation}. Compared with edge-level differential privacy, node-level differential privacy is relatively difficult to be formalized and solve. In~\cite{DBLP:conf/tcc/KasiviswanathanNRS13}, authors contribute several theoretical node-level differential privacy solutions such as Flow-based Lipschitz extension and LP-based Lipschitz extensions. But they all focus on realizing part of the graph properties instead of the graph data itself, such as anonymized degree distribution, subgraph counting, etc. The same kind of research flavor also appeared in relevant node-level differential privacy works like~\cite{DBLP:conf/innovations/BlockiBDS13, DBLP:conf/tcc/KasiviswanathanNRS13, DBLP:conf/focs/RaskhodnikovaS16}. Again, differential privacy mechanisms on graphs is a large and comprehensive topic, a more detailed introduction and extensive literature review can be found in~\cite{DBLP:journals/tkde/JiangPYYGC23}.

\subsection{Other Aspects of Graph Anonymization}
Here, we would also like to review several graph anonymization techniques, but the difference from the majority mentioned above is that: they are not publishing the anonymized graph $G'$ but anonymize some non-trivial and graph statistics of the original graph $G$ and release them to the public~\cite{DBLP:conf/pakdd/WangWW13, DBLP:conf/sigmod/ZhangCPSX15, DBLP:journals/corr/abs-1809-02575, DBLP:conf/nips/UllmanS19}. The central requirement for protecting the graph statistics is that some scalar graph parameters are essential to describe the graph topology (e.g., degree distributions) or even reconstruct the graph topology (e.g., the number of nodes and edge connection probability in the Erdos-Renyi graph). To this end, some methods focus on protecting the important graph parameters and their statistics before releasing them. For example, the spectrum of a graph (i.e., eigen-decomposition of the graph Laplacian matrix) can preserve many important graph properties such as topological connections, low-pass or high-pass graph single filters, etc. Therefore, in~\cite{DBLP:conf/pakdd/WangWW13}, the authors proposed to permute the eigen-decomposition under the differential privacy and then release the permuted parameters. To be specific, given the original eigenvalues and eigenvectors, certain calibrated random noises are sampled and added to them under the differential privacy constraint. Under the same protection mechanism, i.e., differential privacy, the protection goal is set to be the number of occurrences of subgraphs in~\cite{DBLP:conf/sigmod/ZhangCPSX15}, the sequence of degree distribution in directed graphs and undirected graphs in~\cite{DBLP:journals/corr/abs-1809-02575}, and the edge connection probability of random graphs in~\cite{DBLP:conf/nips/UllmanS19}.

\subsection{Challenges and Future Opportunities}
After introducing different graph anonymization techniques, we would like to share some open questions and corresponding challenges.

\subsubsection{Preserving Privacy for Temporal Graphs}
As discussed above, most privacy-preserving graph anonymization methods still consider the input graphs as static. However, in complex real-world scenarios, the graphs are usually evolving over time~\cite{DBLP:conf/kdd/FuZH20, DBLP:conf/kdd/ZhouZ0H20, DBLP:conf/sigir/FuH21, DBLP:conf/bigdataconf/FuH22}, which brings critical challenges to the current privacy-preserving static graph generation process. In other words, the time domain enriches the node attribute dimension and may also dictate the attribute distribution, which leads to increased exposure risk. For example, some graphs contain multiple dynamics and accurately representing them could contribute to graph tasks like classification~\cite{DBLP:journals/fdata/ZhouZXH19, DBLP:conf/kdd/FuFMTH22}. But, the existence of various dynamics increases the probability of being unique and enlarges the leaking risk.

\subsubsection{Preserving Privacy for Heterogeneous Graphs}
During the node identity disclosure and link re-identification, it can be observed that the majority of background knowledge is solely from structural queries, which is already forceful enough. In heterogeneous graphs~\cite{DBLP:journals/sigkdd/SunH12, DBLP:conf/cikm/FuXLTH20}, the abundant node and edge features increase the risk of leaking sensitive information and bring challenges to protection mechanisms, especially the heterogeneous graphs start to evolve~\cite{DBLP:conf/www/Fu0MCBH23, DBLP:conf/www/LiFH23}.\\

To the best of our knowledge, how to generate privacy-preserving heterogeneous or temporal graphs remains open. 
\begin{itemize}
    \item What kind of feature information is sensitive in heterogeneous or time-evolving graphs and should be hidden in the generated graph?
    \item If the corresponding sensitive information is determined, what techniques are effective for protecting structures and features in the heterogeneous or time-evolving environment?
    \item Last but not least, if the corresponding protection mechanism is designed, how to maintain the generation utility simultaneously with privacy constraints?
\end{itemize}

%% file: graph_FL.tex
In recent years, graph machine learning has become increasingly popular due to the abundance of graph-structured data in various domains, such as social networks, recommendation systems, and bioinformatics. However, graph data is usually distributed in multiple data sources, and each data owner does not have enough data to train satisfactory machine learning models, which require a massive amount of graph data. For example, biochemical industries may wish to collaboratively train a graph neural network model to predict the property of molecules. While we introduce one solution with privacy-preserving graph data generation in the last section, another solution is to enable multi-party computation without exchanging raw data. In this section, we introduce federated learning (FL) \cite{fedavg}, a machine learning system where multiple clients (i.e., data owners) collaboratively train machine learning models without exchanging their raw data. In particular, we first introduce the framework of federated learning and its applications with graph data in Subsection \ref{subsec:fl_framework}. Then we introduce important FL algorithms under three representative graph federated learning scenarios: graph-level FL (Subsection \ref{subsec:fl_graph}), subgraph-level FL (Subsection \ref{subsec:fl_subgraph}), and node-level FL (Subsection \ref{subsec:fl_node}). Finally, we summarize the challenges of future opportunities of graph FL in Section \ref{subsec:fl_challenge}. 

\subsection{Framework and Applications of Federated Learning} \label{subsec:fl_framework}

Federated learning (FL) \cite{fedavg} is a distributed learning system where multiple clients (i.e., data sources) collaborate to train a machine learning model under the orchestration of the central server (i.e., the service provider), while keeping their data decentralized and private \cite{advance}. This subsection provides an exposition on the FL framework, followed by an overview of the application of federated learning on graph data. 

\subsubsection{Federated Learning Framework}

A typical FL framework has one central server and $N$ clients, each with its own dataset $\mathcal{D}_i$. The main steps can be summarized as follows: 
\begin{enumerate}
    \item \textit{Parameter broadcasting}. The server broadcasts the current global model to (selected) clients. 
    \item \textit{Local update}. Each client locally trains its local model. 
    \item \textit{Parameter uploading}. Each client sends upload the model update back to the server. 
    \item \textit{Model aggregation}. The server aggregates the model updates collected from clients and updates the global model. 
    \item \textit{Repeat}: Steps 1-4 are repeated for multiple communication rounds until the global model converges to satisfactory performance. 
\end{enumerate}

One of the most popular FL algorithms is FedAvg \cite{fedavg}. In each communication rounds, the server randomly selects a subset of clients, and broadcasts the global model to them. Each client locally updates the model with multiple iterations of stochastic gradient descent, and uploads its local model back to the server. Finally, the server computes a weighted average of local model parameters, and updates the global model parameters. Algorithm \ref{alg:fedavg} gives the pseudo-code of FedAvg. Notice that in FedAvg, local data never leaves the client side. Besides FedAvg, most of the FL algorithms strictly follow the aforementioned training protocol \cite{fedprox,per-fedavg}, or roughly follow it with a few modifications \cite{ifca,fedmix}. 

\input{Blocks/fedavg}

FL protects client privacy in two main ways. Firstly, instead of transmitting the raw data, FL transmits only the model parameters, which are updated based on the local data of each client. By doing so, FL ensures that sensitive data remains on the client's device and is not transmitted to the central server and other clients. Secondly, the model parameters uploaded to the server only reveal the distribution of local data, rather than individual data points. This approach helps to maintain privacy by obscuring the specific data points used to train the model. 

FL can be equipped with differential privacy mechanisms \cite{dp0,dp1} to enhance privacy protection. As described in the last section, differential privacy is a technique that involves adding noise to data in order to obscure individual contributions while still maintaining overall data patterns. However, different from graph generation, where the noise is added to the data (e.g., node feature, edges, etc), in the context of FL, the noise is added to the uploaded and downloaded model parameters. This ensures that even if an attacker were to obtain the model parameters, they would not be able to accurately infer the raw data from the model parameter. By adding moderate noise to the parameters, the model's accuracy may be slightly reduced, but the overall performance remains comparable to non-private models. In summary, by using differential privacy mechanisms, FL can achieve even better privacy protection by making it harder for attackers to identify the sensitive data contributed by individual clients. 


\subsubsection{Application of Graph Federated Learning} 

In this part, we introduce important applications of federated learning on graph data. Roughly, we survey three representative application scenarios: \textit{graph-level FL}, \textit{subgraph-level FL}, and \textit{node-level FL}. 

\begin{enumerate}
    \item \textit{Graph-level FL}: Each client has one or several graphs, while different graphs are isolated and independent. One typical application of graph-level FL is for drug discovery \cite{spreadgnn}, where biochemical industries collaborate to train a graph neural network model predicting the property of molecules. Each molecule is a graph with basic atoms as nodes and chemical bonds as edges. 
    
    \item \textit{Subgraph-level FL}: Each client has one graph, while each graph is a subgraph of an underlying global graph. One representative application of subgraph-level FL is for financial transaction data \cite{fedcog}. Each FL client is a bank that keeps a graph encoding the information of its customers, where nodes are individual customers and edges are financial transaction records. While each bank holds its own graph, customers in one bank may have connections to customers in another bank, introducing cross-client edges. Thus, each bank's own graph is a subgraph of an underlying global graph. 
    
    \item \textit{Node-level FL}: Each client is a node of a graph, and edges are the pairwise relationships between clients, e.g., their distribution similarity or data dependency. One example is the smart city, where clients are traffic sensors deployed on the road and linked to geographically adjacent sensors. While clients form a graph, each client can make an intelligent decision based on the collected road conditions and nearby devices. 
    
\end{enumerate}

Figure \ref{fig:fl} illustrates the three application scenarios above. Next, we investigate each application scenario in the following three subsections individually. 

\input{Blocks/fl_figure}




\subsection{Graph-level FL} \label{subsec:fl_graph}

In this subsection, we investigate graph-level FL. Graph-level FL is a natural extension of traditional FL: while each client has one or several graphs, different graphs are isolated and independent. The goal of each client is to train a graph neural network (GNN) model for a variety of local tasks, e.g., node-level (e.g., node classification), link-level (e.g., edge prediction), or graph-level (e.g., graph classification). 

One of the most representative applications of graph-level FL is drug discovery, where graphs are molecules with atoms as nodes and chemical bonds as edges. Each FL client can be a pharmaceutical corporation that owns molecule data. Multiple corporations collaborate to train better model for molecular property prediction. 

The biggest challenge of graph-level FL is the non-identical distribution among different clients' data. Since each client in FL collects their local data individually, their local datasets usually have a different distribution. For example, different pharmaceutical corporations may focus on different types of molecules. Such heterogeneity among clients' data distributions introduces optimization challenges to FL. Moreover, when clients' distribution is largely different, it might be harmful or even impossible to train one universal global model across all clients. More sophisticated techniques are required to achieve beneficial collaboration. 

Next, we will introduce algorithms for graph-level FL in two parts: global federated learning and personalized federated learning. Since graph-level FL is a natural extension of traditional FL, we will cover both general FL algorithms and graph FL algorithms. 


\subsubsection{Global Federated Learning}

Global federated learning (GFL) aims to train a \textit{shared} global model for all clients. FedAvg \cite{fedavg} provides an initial solution for training GNNs with isolated graphs from multiple clients. However, when clients have significantly different underlying distributions, FedAvg needs much more communication rounds for convergence to a satisfactory model, and may converge to a sub-optimal solution \cite{fedavg}. This phenomenon of worse convergence is usually explained by \textit{weight divergence} \cite{data-share}, i.e, even with the same parameter initialization, the model parameters for different clients are substantially different after the first local stochastic gradient descent (SGD) step. With different model parameters, the mean of client gradients can be different from the gradient in centralized SGD, and introduce error to the model loss \cite{fedopt}. 

\textbf{Data-sharing}. 
To tackle the non-IID challenge to FL optimization, a simple but effective method is to share a small amount of data among clients. \cite{data-share} first explore an association between the weight divergence and the non-IIDness of the data, and propose a method to share a small amount of data among the server and all clients. As a result, the accuracy can be increased by ~30\% for the CIFAR-10 dataset \cite{cifar} with only 5\% globally shared data. \cite{fedmix} further improves the privacy of this approach by sharing the average of local data points, instead of raw data. Specifically, each client uploads averaged data, receives averaged data from other clients, and performs Mixup \cite{mixup} data augmentation locally to alleviate weight divergence. However, both methods require modification of the standard FL protocol and transmission of data. Another way to improve privacy is to share synthetic data generated by generative adversarial networks (GANs) \cite{gan}, instead of the raw data. The synthetic data can be a collection of each client's synthetic data generated with local GANs or generated with one global GAN trained in FL \cite{fedgan,effgan}. However, it is unclear whether GAN can provide enough privacy, since it may memorize the training data \cite{gan-theory}. 

\textbf{Modifying local update}. 
Another line of research works modifies the local update procedure to alleviate weight divergence without changing the communication protocol of FL. FedProx \cite{fedprox} adds a proximal term to the local objective to stabilize the training procedure. The proximal term is the squared L2 distance between the current global model and the local model, which prevents the local model from drifting too far from the global model. SCAFFOLD \cite{scaffold} estimates how local updates deviate from the global update, and it then corrects the local updates via variance reduction. Based on the intuition that the global model can learn better representation than local models, MOON \cite{moon} conducts contrastive learning at the model level, encouraging the agreement of representation learned by the local and global models. 


\subsubsection{Personalized Federated Learning}

While the aforementioned algorithms can accelerate the model optimization for GFL, one model may not always be ideal for all participating clients \cite{cfl}. Recently, personalized federated learning (PFL) has been proposed to tackle this challenge. PFL allows FL clients to collaboratively train machine learning models while each client can have different model parameters. 

\textbf{Clustered FL}. 
In clustered FL, clients are partitioned into non-overlapping groups. Clients in the same group will share the same model, while clients from different groups can have different model parameters. In IFCA \cite{ifca}, $k$ models are initialized and transmitted to all clients in each communication round, and each client picks the model with the smallest loss value to optimize. FedCluster \cite{cfl} iteratively bipartition the clients based on their cosine similarity of gradients. GCFL \cite{gcfl} generalizes this idea to graph data, enabling collaborative training with graphs from different domains. Observing that the gradients of GNNs can be fluctuating, GCFL+ \cite{gcfl} uses a gradient sequence-based clustering mechanism to form more robust clusters. 

\textbf{Personalized Modules}. 
Another prevalent way for PFL is personalized modules. In these works, the machine learning model is divided into two parts: the shared part and the personalized part. The key is to design a model structure suitable for personalization. For example, when a model is split into a feature extractor and classifier, FedPer \cite{fedper} shares the feature extractor and personalizes the classifier, while LG-FedAvg \cite{lg-fedavg} personalizes the feature extractor and shares the classifier. Similar techniques in used in FMTGL \cite{fmtgl} and NGL-FedRep \cite{ngl-fedrep}. Moreover, PartialFed \cite{partialfed} can automatically select which layers to personalize and which layers to share. On graph data, \cite{fedstar} observe that while the feature information can be very different, some structural properties are shared by various domains, revealing the great potential for sharing structural information in FL. Inspired by this, they propose FedStar that trains a feature-structure decoupled GNN. The structural encoder is globally shared across all clients, while the feature-based knowledge is personalized. 

\textbf{Local Finetuning and Meta-Learning}. 
Finetuning is widely used for PFL. In these works, a global model is first trained with all clients. The global model encodes the information of the population but may not adapt to each client's own distribution. Therefore, each client locally finetunes the global model with a few steps of gradient descent. Besides vanilla finetuning, Per-FedAvg \cite{per-fedavg} combines FL with MAML \cite{maml}, an algorithm for meta-learning, to improve the performance of finetuning. Similarly, pFedMe \cite{pfedme} utilize Moreau Envelopes for personalization. It adds a proximal term to the local finetuning objective, and aims to find a local model near the global model, with just a few steps of gradient descent. GraphFL \cite{graphfl} applies a similar meta-learning framework on graph data, addressing the heterogeneity among graph data and handling new label domains with a few new labeled nodes. 

\textbf{Multi-task Learning}. 
PFL is also studied within the framework of multi-task learning. MOCHA \cite{mocha} uses a matrix to model the similarity among each pair of clients. Clients with similar distribution will be encouraged to have similar model parameters. FedGMTL \cite{spreadgnn} generalizes this idea to graph data. Similarly, SemiGraphFL \cite{semigraphfl} computes pairwise cosine similarity among clients' hidden representations. As a result, clients with more similar data will have greater mutual influence. However, it requires the transmission of hidden representation. FedEM \cite{fedem} assumes that each client's distribution is a mixture of unknown underlying distributions and proposes FedEM, an EM-like algorithm for multi-task FL. Finally, FedFOMO \cite{fedfomo} allows each client to have a different mixture weight of local models during the aggregation steps. It provides a flexible way for model aggregation. 

\textbf{Graph Structure Augmentation}. 
In the previous works, graph structures are considered as ground truth. However, graphs can be noisy or incomplete, which can hurt the performance of GNNs. To tackle incomplete graph structures, FedGSL \cite{fedgsl} optimizes the local client's graph and GNN parameters simultaneously. 










\input{Blocks/repository}



\subsection{Subgraph-level FL} \label{subsec:fl_subgraph}

Similar to graph-level FL, each client in subgraph-level FL holds one graph. However, clients' graphs are a subgraph of a latent large entire graph. In other words, there are cross-client edges in the entire graph, where the two nodes of these edges belong to different clients. The task is usually node-level, while the cross-client edges can contribute to the task. 

One application of subgraph-level FL is financial fraud detection. Each FL client is a bank aiming to detect potential fraud with transaction data. Each bank keeps a graph of the information of its customers, where nodes are individual customers and edges are transaction records. While each bank holds its own graph, customers in one bank may have connections to customers in another bank, introducing edges across clients. These cross-client edges help to train better ML models. 

The biggest challenge for subgraph-level FL is to handle cross-client edges. In GNNs, each node iteratively aggregates information from its neighboring nodes, which may be from other clients. However, during local updates in traditional FL, clients cannot get access to the data from other clients. Directly exchanging raw data among clients is prohibited due to privacy concerns. It is challenging to enable cross-client information exchange while preserving privacy. Moreover, when nodes' identities are not shared across clients, the cross-client edges can be missing and stored in none of the clients. Even if we collect clients' local subgraphs, we cannot reconstruct the global graph. 

In this subsection, we will mainly focus on two scenarios. In the first part, we introduce algorithms when the hidden entire graph is given but stored separately in different clients. In the second part, we consider a more challenging setting: the cross-client edges are missing, and we cannot simply concatenate local graphs to reconstruct the entire graph losslessly. We focus on how to generate these missing edges or missing neighbors for each node. 


\subsubsection{Cross-client Propagation}

When the cross-client edges are available, the major challenge is to enable cross-client information propagation without leaking raw data. FedGraph \cite{fedgraph} designs a novel cross-client convolution operation to avoid sharing raw data across clients. It avoids exchanging representations in the first GCN layer. Similarly, FedPNS \cite{fedpns} control the number of neighbor sampling to reduce communication costs. 
FedCog \cite{fedcog} proposes graph decoupling operation, splitting local graph to internal graph and border graph. The graph convolution is accordingly divided into two sequential steps: internal propagation and border propagation. In this process, each client sends the intermediate representation of internal nodes to other clients. 
Considering that directly exchanging feature representations between clients can leak private information. In user-item graphs, FedPerGNN \cite{fedpergnn} design a privacy-preserving user-item graph expansion protocol. Clients upload encrypted item IDs to the trusted server, and the server matches the ciphertexts of item IDs to find clients with overlapping item IDs. DP-FedRec \cite{fedrec} uses private set intersection to exchange the edges information between clients and applies differential privacy techniques to further protect privacy. 
Different from the above methods, FedGCN \cite{fedgcn} does not rely on communication between clients. Instead, it transmits all the information needed to train a GCN between the server and each client, only once before the training. Moreover, each node at a given client only needs to know the accumulated information about the node's neighbors, which reduces possible privacy leakage. 


\subsubsection{Missing Neighbors}

For some applications, the cross-client edges can be missing or not stored in any clients. Notice that although each client also holds a disjoint graph in graph-level FL, graph-level FL and subgraph-level FL with missing neighbors are substantially different. For graph-level FL, there are essentially no cross-client edges. For example, there are no chemical bonds between two molecules from different corporations' datasets. However, for subgraph-level FL, the cross-client edges exist, but are missing in certain applications. We may get suboptimal GNN models if ignoring the existence of cross-client edges. Therefore, the major challenge is to reconstruct these missing edges, or reconstruct missing neighbors for each node. 

FedSAGE \cite{fedsage} first defines the missing neighbors' challenge, and proposes a method the generate pseudo neighbors for each node. It uses existing subgraphs to train a neighbors generator and generate one-hop neighbors for each client to mend the graph. Since missing neighbors are generated locally, no feature exchange is required between clients after the local subgraphs are mended. However, the training of neighbor generators requires cross-client hidden representation exchanges. Similarly, FedNI \cite{fedni} uses a graph GAN model to generate missing nodes and edges. 





\subsection{Node-level FL} \label{subsec:fl_node}

The final application scenario of graph federated learning is node-level. Different from the aforementioned two scenarios, each client in node-level FL can hold any type of data, not restricted to graphs. Instead, the clients themselves are nodes in a graph, while the edges are their pairwise relationship of communication or distribution similarity. 

One typical application of node-level FL is the Internet of Things (IOT) devices in a smart building \cite{g-fedfilt}. Due to bandwidth constraints, it can be costly for each IoT device to communicate with the central server. However, IoT devices in the same local area network can communicate very efficiently. As a result, IoT devices form a graph with pairwise communication availability as edges. Another application is for the smart city \cite{fedgs}, where clients are traffic sensors deployed on the road and linked to geographically adjacent sensors. Each device can collect data and make the real-time decision without waiting for the response of cloud servers. Each device needs to make an intelligent decision based on the collected road conditions and nearby devices. 

In this subsection, we will first introduce algorithms where the graph models communication constraints among clients. In these works, there is no central server, and clients can only exchange information along edges. Then, we will introduce algorithms where the graph models the relationship between clients' distributions. In these works, although a central server is available, the graph among clients models distributional similarity or dependency among clients, potentially contributes to the model performance. 


\subsubsection{Graph as Communication Network}

Traditional FL relies on a central server to enable communication among clients. Each client trusts the central server and uploads their model update to the server. However, in many scenarios, a trusted central server may not exist. Even when a central server exists, it may be expensive for clients to communicate with the server. Therefore, serverless FL (a.k.a. peer-to-peer FL) has been studied to relieve communication constraints. 

The standard solution for serverless FL is fully decentralized FL \cite{decentral,p2p}, where each client only averages its model parameter with its neighbors. D-FedGNN \cite{d-fedgnn} uses these techniques to train GNN models. SpreadGNN \cite{spreadgnn} generalizes this framework to personalized FL, where each client has non-IID data and a different label space.

\subsubsection{Graph as Distribution Similarities}

When the central server is available, a graph of clients may still be beneficial when it models distributional relationships among clients. When edges link clients with highly similar distributions, parameter sharing along edges can potentially improve the model performance for both clients. When edges link clients with data dependency, information exchange along edges can even provide additional features for inference. 

FedGS \cite{fedgs} models the data correlations of clients with a data-distribution-dependency graph, and improves the unbiasedness of the client sampling process. Meanwhile, SFL \cite{sfl} assumes a pre-defined client relation graph stored on the server, and the client-centric model aggregation is conducted along the relation graph’s structure. GraphFL \cite{graphfl2} considers client-side information to encourage similar clients to have similar models. BiG-Fed \cite{big-fed} applies graph convolution on the client graph, so each client's prediction can benefit from its neighbors with highly correlated data. Finally, \cite{d2d-fedl} designs a client sampling technique considers both communication cost and distribution similarity.







Finally, we summarize the official implementation of FL algorithms and useful repositories in Table \ref{tab:fl_repository}.

\subsection{Challenges and Future Opportunities} \label{subsec:fl_challenge}

In this part, we present several limitations in current works and provide open problems for future research. 

\subsubsection{Model Heterogeneity for Graph-Level FL}
In previous works of graph-level FL, although each FL client usually has different data distribution it is usually assumed that the model architecture is shared across all clients. However, the optimal architecture for different clients can be different. For example, a well-known issue in GNNs is the over-smoothing problem. When the number of graph convolutional layers is higher than the diameter of the graph, GNN models may learn similar representations for all nodes in the graph, which harms the model performance. When each FL clients hold a substantially different size of graphs, it is highly likely that the optimal depth of the GNN model is different for them. 

\subsubsection{Avoiding Cross-Client Transmission for Sub-graph-Level FL}
Most of the previous subgraph-level FL algorithms highly rely on direct information exchange along cross-client edges. While such operations are natural variants of graph convolution, such operations also raise privacy concerns. Moreover, different from traditional FL where each client downloads aggregated model parameters that reveal the population, feature exchange along the edges can expose information about individuals. It would be beneficial if the cross-client transmission can be avoided without greatly degrading the model.

%% file: Blocks/fedavg.tex
\newcommand{\SUB}[1]{\ENSURE \hspace{-0.15in} \textbf{#1}}
\newcommand{\clientfrac}{\ensuremath{C}}
\newcommand{\pp}{\mathcal{P}}
\newcommand{\lbs}{\ensuremath{B}}  
\newcommand{\lepochs}{\ensuremath{E}} 
\newcommand{\grad}{\triangledown}
\newcommand{\loss}{\ell}
\newcommand{\nc}{K}
\newcommand{\mycaptionof}[2]{\captionof{#1}{#2}}
\renewcommand{\algorithmicensure}{}

\begin{algorithm}[t]
\begin{algorithmic}[1]
\SUB{Server executes:}
   \STATE initialize model parameter $w_0$
   \FOR{each round $t = 1, 2, \dots, T$}
     \STATE $m \leftarrow \max(\clientfrac\cdot K, 1)$
     \STATE $S_t \leftarrow$ (random set of $m$ clients)
     \FOR{each client $k \in S_t$ \textbf{in parallel}}
       \STATE $w_{t+1}^k \leftarrow \text{ClientUpdate}(k, w_t)$ 
     \ENDFOR
     \STATE $m_t \leftarrow \sum_{k \in S_t} n_k$
     \STATE $w_{t+1} \leftarrow \sum_{k \in S_t} \frac{n_k}{m_t} w_{t+1}^k$
   \ENDFOR
   \STATE

 \SUB{ClientUpdate($k, w$):}\ \ \  // \emph{Run on client $k$}
  \STATE $\mathcal{B} \leftarrow$ (split $\pp_k$ into batches of size $\lbs$)
  \FOR{each local epoch $i$ from $1$ to $\lepochs$}
    \FOR{batch $b \in \mathcal{B}$}
      \STATE $w \leftarrow w - \eta \grad \loss(w; b)$
    \ENDFOR
 \ENDFOR
 \STATE return $w$ to server
\end{algorithmic}
\mycaptionof{algorithm}{FedAvg. The $\nc$
  clients are indexed by $k$; $C$ is the participation rate, $\lbs$ is the local minibatch size,
  $\lepochs$ is the number of local epochs, and $\eta$ is the learning
  rate.}\label{alg:fedavg}
\end{algorithm}

%% file: Blocks/fl_figure.tex
\begin{figure*}[h!]
    \centering%
    \subcaptionbox{Graph-level (Subsection 3.2)} 
    {\includegraphics[width=0.3\linewidth]{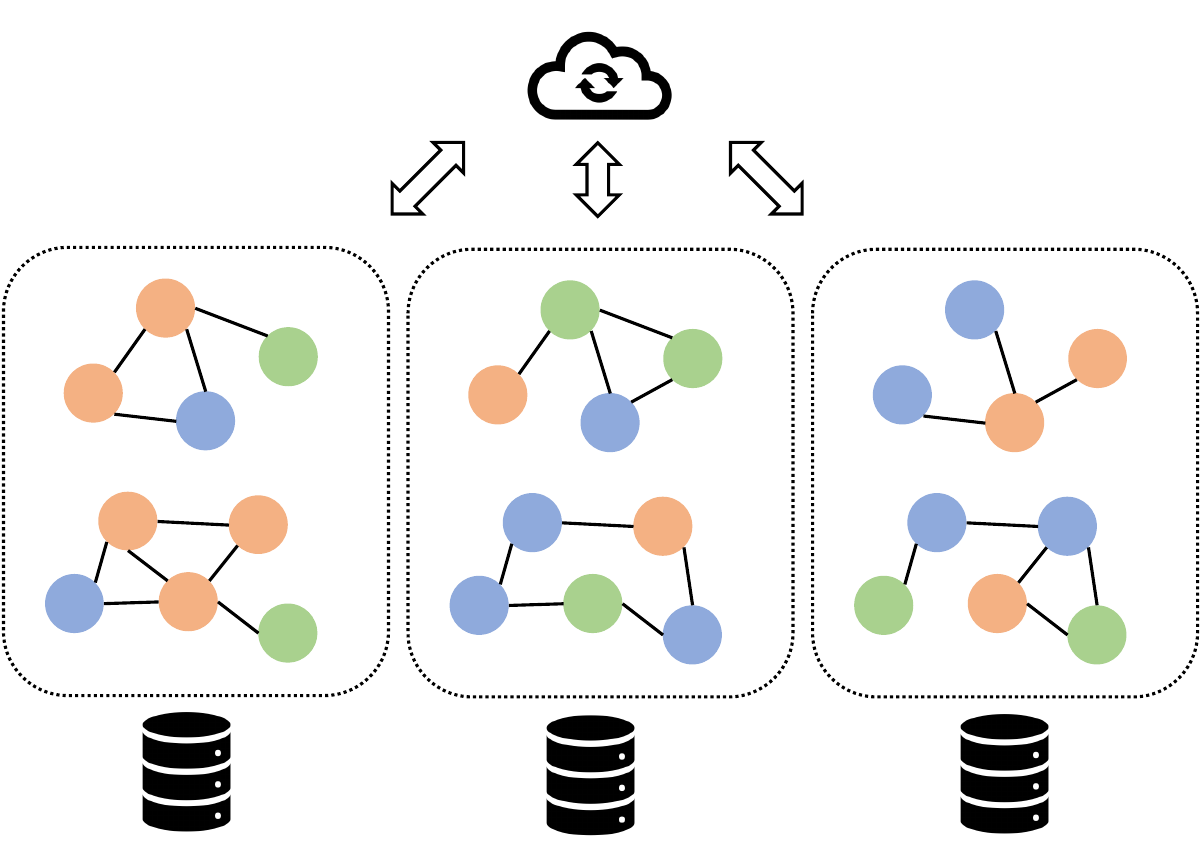}}
    \hskip0.5cm
    \subcaptionbox{Subgraph-level (Subsection 3.3)} 
    {\includegraphics[width=0.3\linewidth]{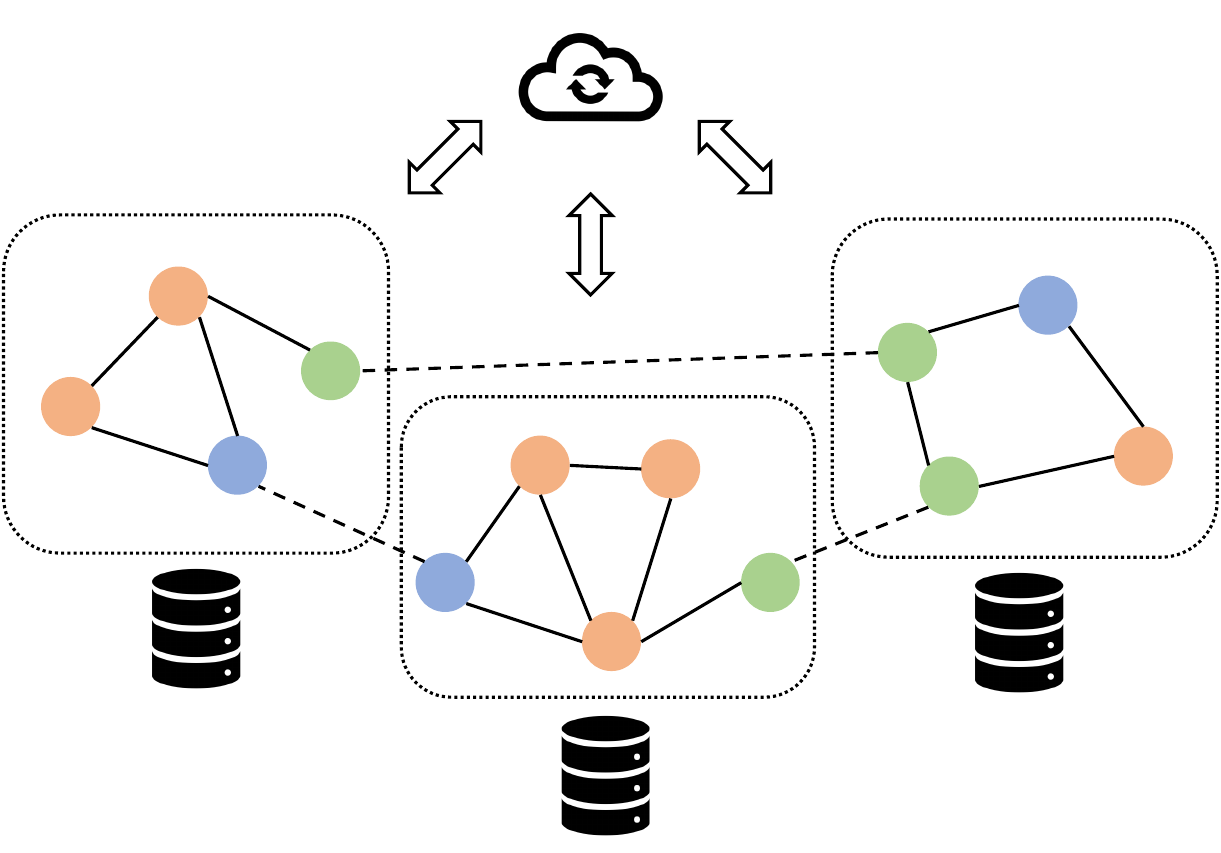}}
    \hskip0.5cm
    \subcaptionbox{Node-level (Subsection 3.4)} 
    {\includegraphics[width=0.3\linewidth]{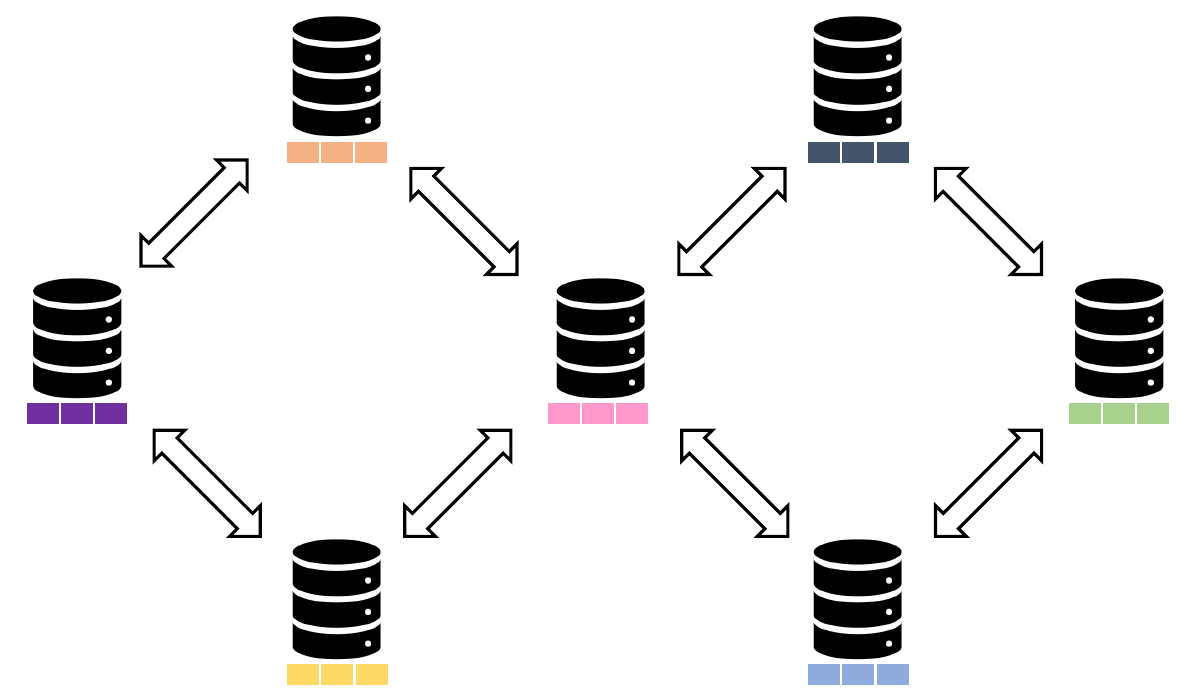}}
    \caption{Three application scenarios of graph federated learning. }
    \label{fig:fl}
\end{figure*}

%% file: Blocks/repository.tex
\begin{table*}[t]
\caption{Repositories for Graph Federated Learning \label{tab:fl_repository}}
\centering
\footnotesize
\begin{tabular}{llp{9cm}l}
\toprule 
Scenario & Name & Description & Link \\
\midrule
\multirow{8}{*}{Graph-level} 
& FedProx \cite{fedprox} &  A general GFL algorithm with modified local update & \href{https://github.com/litian96/FedProx}{Github}  \\
& IFCA \cite{ifca} &        A general clustered FL algorithm & \href{https://github.com/jichan3751/ifca}{Github} \\
& GCFL \cite{gcfl} &        A graph-specific clustered FL algorithm & \href{https://github.com/Oxfordblue7/GCFL}{Github} \\
& LG-FedAvg \cite{lg-fedavg} & A general PFL algorithm with personalized modules & \href{https://github.com/pliang279/LG-FedAvg}{Github} \\
& FedStar \cite{fedstar} &  A graph-specific PFL algorithm with personalized modules & \href{https://github.com/yuetan031/FedStar}{Github} \\
& pFedMe \cite{pfedme} &    A general PFL algorithm based on meta-learning & \href{https://github.com/CharlieDinh/pFedMe}{Github} \\
& GraphFL \cite{graphfl} &  A graph-specific PFL algorithm based on meta-learning & \href{https://github.com/binghuiwang/GraphFL}{Github} \\
& FedFOMO \cite{fedfomo} &                 A general PFL algorithm based on multi-task learning & \href{https://github.com/NVlabs/FedFomo}{Github}\\

\midrule

\multirow{2}{*}{Subgraph-level} 
& FedGCN \cite{fedgcn} &    An FL algorithm with one-shot cross-client propagation & \href{https://github.com/yh-yao/FedGCN}{Github} \\
& FedSAGE \cite{fedsage} &  An FL algorithm with missing neighbors generation & \href{https://github.com/zkhku/fedsage}{Github} \\

\midrule

\multirow{3}{*}{Node-level} 
& SpreadGNN \cite{spreadgnn} &  A serverless PFL algorithm  & \href{https://github.com/FedML-AI/SpreadGNN}{Github} \\
& FedGS \cite{fedgs} &      An FL algorithm with graph as distribution similarities & \href{https://github.com/WwZzz/FedGS}{Github} \\
& SFL \cite{sfl} &          An GL with pre-defined graph for server aggregation & \href{https://github.com/dawenzi098/SFL-Structural-Federated-Learning}{Github} \\

\midrule

\multirow{7}{*}{Others} 
& TensorFlow Federated & A framework for implementing federated learning & \href{https://github.com/tensorflow/federated}{Github} \\
& FedLab \cite{fedlab} & A Flexible Federated Learning Framework & \href{https://github.com/SMILELab-FL/FedLab}{Github} \\
& PFL-Non-IID & Reproduction of popular PFL algorithms & \href{https://github.com/TsingZ0/PFL-Non-IID}{Github}\\
& FedGraphNN \cite{fedgraphnn} & FedGraphNN: A Federated Learning System and Benchmark for Graph Neural Networks & \href{https://github.com/FedML-AI/FedML/tree/master/python/app/fedgraphnn}{Github} \\
& FederatedScope-GNN \cite{federatedscope-gnn} & A unified, comprehensive and efficient package for Federated Graph Learning  & \href{https://github.com/alibaba/FederatedScope}{Github} \\

\bottomrule
\end{tabular}
\end{table*}

%% file: envision.tex

In this section, we analyze the current developments and limitations of privacy-preserving graph machine learning, and explain the necessity of combining them. In addition, we identify a number of unsolved research directions that could be addressed to improve the privacy of graph machine learning systems. 

\subsection{Limitation of Current Techniques}

In the previous two sections, we introduced privacy-preserving graph data generation and computation, respectively. However, both techniques have their own limitations. 

\begin{itemize}
    \item For privacy-preserving graph generation, while it can provide good privacy protection for graph data, it also has a significant drawback on model utility. The privacy-preserving techniques applied during data generation are not designed for specific machine learning tasks and may influence the utility of the resulting model. For example, consider a graph with four nodes $a$, $b$, $c$, and $d$. The nodes $a$ and $b$ have a positive label, while $c$ and $d$ have a negative label. Switching the edges from $(a, b), (c, d)$ to $(a, c), (b, d)$ does not change the degree distribution of the graph, but it changes the graph from a homophilous graph to a heterophilous graph, i.e., edges are more likely to link two nodes with different labels. This change can harm the performance of many GNN models, which are designed to work well with homogeneous graphs \cite{nonhomo}. It is important to consider the downstream machine learning tasks when designing privacy-preserving techniques for graph data. 

    \item For privacy-preserving graph computation, while FL can avoid the transmission of raw data, it has been shown that transmitting raw model parameters or gradients may not provide enough privacy, as attackers can use the gradient or model update to reconstruct private data \cite{deepleak,inv}. Moreover, many subgraph-level and node-level federated learning algorithms require the transmission of hidden representations, which can also leak private information. Therefore, protecting the raw data from being reconstructed is essential to federated learning systems.
\end{itemize}

\subsection{Combination of Privacy-Preserving Graph Data Generation and Computation}

\begin{figure}[ht]
\vspace{10pt}
\includegraphics[width=0.43\textwidth]{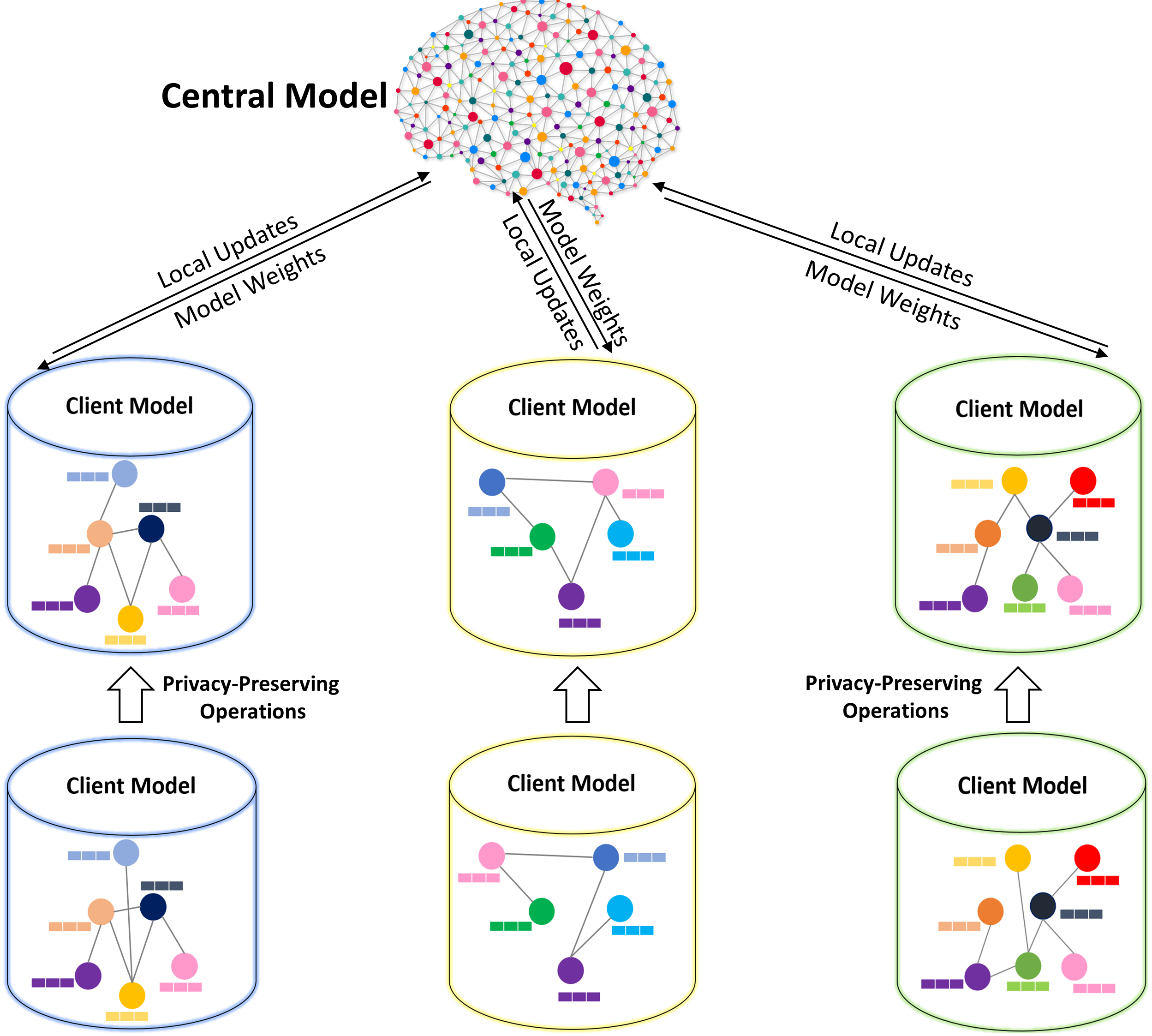}
\centering
\caption{Privacy-preserving Graph Data with Privacy-preserving Computation.}
\label{fig:combination}
\end{figure}

To address the limitations of current privacy-preserving techniques, it is essential to combine privacy graph data generation with the graph federated learning frameworks, as shown in Figure \ref{fig:combination}. This approach can provide an effective solution to the privacy preservation issues of graph machine learning models.

Specifically, the generated synthetic data is used instead of the real data during the training process. This means that even if the transmitted information is decrypted, it is just from the generated synthetic data and not the real data. The synthetic data can be generated in such a way that it preserves the statistical properties of the original data while ensuring privacy preservation. This can be achieved using various techniques, including differential privacy, homomorphic encryption, and secure multi-party computation.

The combination of privacy graph data generation and graph federated learning frameworks has several benefits. First, it ensures privacy preservation during the training process by using synthetic data. Second, it enables the transfer of graph machine learning model parameters rather than embedding vectors or other information. This can improve the accuracy and efficiency of the model. Finally, it provides a robust defense against privacy attacks and reverse-engineering, as the transmitted information is just from the generated synthetic data and not the real data.

\subsection{Future Directions}

Combining privacy-preserving data generation and computation is a promising approach to protect individual privacy while maintaining model utility in machine learning. However, it also poses several challenges and possible future directions. 

\subsubsection{Distribution of Privacy Budget}

When combining privacy-preserving data generation with computation, noises are added to both raw data and model parameters. However, it is still unclear how to distribute the privacy budget between data generation and computation in a way that optimizes the privacy-utility trade-off. In this approach, noises are added to the graph data during data generation and to the model parameters during data computation (i.e., federated learning), which results in an overall reduction in accuracy. However, while the privacy analysis for data generation is directly defined on the data space, the privacy analysis for federated learning requires transforming the change on parameter space back to data space. Such transformation requires estimating the sensitivity of a machine learning algorithm (i.e., how the change of a data point affects the learned parameters), which is only loosely bounded in current works \cite{dp0,dp1}. A more precise analysis of privacy is required to better understand the impact of privacy budget allocation on the overall privacy-utility trade-off. 

\subsubsection{Parameter Information Disentanglement}

Another future challenge when combining privacy-preserving data generation and computation is the disentanglement of task-relevant and task-irrelevant information. Currently, the noise added to the model parameters is isotropic, meaning that task-relevant and task-irrelevant information are equally protected. However, not all information is equally important for model utility. If we can identify which information has a significant influence on model performance, we can distribute more privacy budget to this information while allocating less privacy budget to task-irrelevant information. This can result in a better privacy-utility trade-off. Disentangling task-relevant and task-irrelevant information would require a more sophisticated analysis of model architecture and data characteristics to determine which features contribute most to model performance.

%% file: conclusion.tex
In this paper, we review the research for privacy-preserving techniques for graph machine learning from the data to the computation, considering the situation where the data need to be shared or are banned from being transmitted. To be specific, for privacy-preserving graph data generation techniques, we analyze the forceful attackers first and then introduce how corresponding protection methods are proposed to defend attackers. For the privacy graph data computation, we circle around the federated learning setting and discuss how the general federated learning framework applied to graph data and what the potential challenges originated from non-IIDness, and how the nascent research works address them. In the end, we analyze the current limitation and propose several promising research directions.